\def\year{2020}\relax
\documentclass[letterpaper]{article} 
\usepackage{aaai20}  
\usepackage{times}  
\usepackage{helvet} 
\usepackage{courier}  
\usepackage[hyphens]{url}  
\usepackage{graphicx} 
\usepackage{graphicx}  
\usepackage[symbol]{footmisc}

\newtheorem{definition}{Definition}

\usepackage[linesnumbered,ruled,vlined]{algorithm2e}
\usepackage{graphicx} 
\usepackage{subfig}
\usepackage{verbatim}
\usepackage{booktabs}
\usepackage{amsfonts}

\title{Multi-Agent Game Abstraction via Graph Attention Neural Network}

\author{
Yong Liu, \textsuperscript{\rm 1}\thanks{Equal contribution, {\dag} corresponding author. This work is partially done when Weixun Wang was intern in NetEase Fuxi AI Lab.}
Weixun Wang, \textsuperscript{\rm 2}\footnotemark[1]
Yujing Hu, \textsuperscript{\rm 3}
Jianye Hao, \textsuperscript{\rm 2,4}\footnotemark[2]
Xingguo Chen, \textsuperscript{\rm 5} 
Yang Gao\textsuperscript{\rm 1}\footnotemark[2]\\
\textsuperscript{\rm 1}{National Key Laboratory for Novel Software Technology, Nanjing University}\\
\textsuperscript{\rm 2}{Tianjin University}, 
\textsuperscript{\rm 3}{NetEase Fuxi AI Lab},
\textsuperscript{\rm 4}{Noah's Ark Lab, Huawei}\\
\textsuperscript{\rm 5}{Jiangsu Key Laboratory of Big Data Security \& Intelligent Processing,}\\
{Nanjing University of Posts and Telecommunications}\\
lucasliunju@gmail.com, \{wxwang, jianye.hao\}@tju.edu.cn \\
huyujing@corp.netease.com, chenxg@njupt.edu.cn, gaoy@nju.edu.cn\\
}

\begin{document}

\maketitle

\begin{abstract}


In large-scale multi-agent systems, the large number of agents and complex game relationship cause great difficulty for policy learning. Therefore, simplifying the learning process is an important research issue. In many multi-agent systems, the interactions between agents often happen locally, which means that agents neither need to coordinate with all other agents nor need to coordinate with others all the time. Traditional methods attempt to use pre-defined rules to capture the interaction relationship between agents. However, the methods cannot be directly used in a large-scale environment due to the difficulty of transforming the complex interactions between agents into rules. In this paper, we model the relationship between agents by a complete graph and propose a novel game abstraction mechanism based on two-stage attention network (G2ANet), which can indicate whether there is an interaction between two agents and the importance of the interaction. We integrate this detection mechanism into graph neural network-based multi-agent reinforcement learning for conducting game abstraction and propose two novel learning algorithms GA-Comm and GA-AC. We conduct experiments in Traffic Junction and Predator-Prey. The results indicate that the proposed methods can simplify the learning process and meanwhile get better asymptotic performance compared with state-of-the-art algorithms.

\end{abstract}

\section{Introduction}


Multi-agent reinforcement learning (MARL) has shown a great success for solving sequential decision-making problems with multiple agents. Recently, with the advance of deep reinforcement learning (DRL) \cite{mnih2016asynchronous,schulman2017proximal}, the combination of deep learning and multi-agent reinforcement learning has also been widely studied \cite{foerster2018counterfactual,sunehag2018value,rashid2018qmix}.


Recent work has focused on multi-agent reinforcement learning in large-scale multi-agent systems \cite{yang2018mean,chen2018factorized}, in which the large number of agents and the complexity of interactions pose a significant challenge to the policy learning process. Therefore, simplifying the learning process is a crucial research area.
Earlier work focuses on loosely coupled multi-agent systems, 
and adopt techniques such as game abstraction and knowledge transfer to help with speeding up multi-agent reinforcement learning\cite{GuestrinLP02,KokV04,de2010learning,MeloV11,hu2015learning,yu2015multiagent,yong2019ijcai}. However, in a large multi-agent environment, agents are often related to some other agents rather than independent, which makes the previously learnt single-agent knowledge has limited use. Recent work focuses on achieving game abstraction through pre-defined rules (e.g., the distance between agents) \cite{yang2018mean,jiang2018graph}. However, it is difficult to define the interaction relationship between agents through pre-defined rules in large-scale multi-agent systems. In this paper, we propose to automatically learn the interaction relationship between agents through end-to-end model design, based on which game abstraction can be achieved.

\begin{figure}[tbp]
\centering
\includegraphics[width=3.3in]{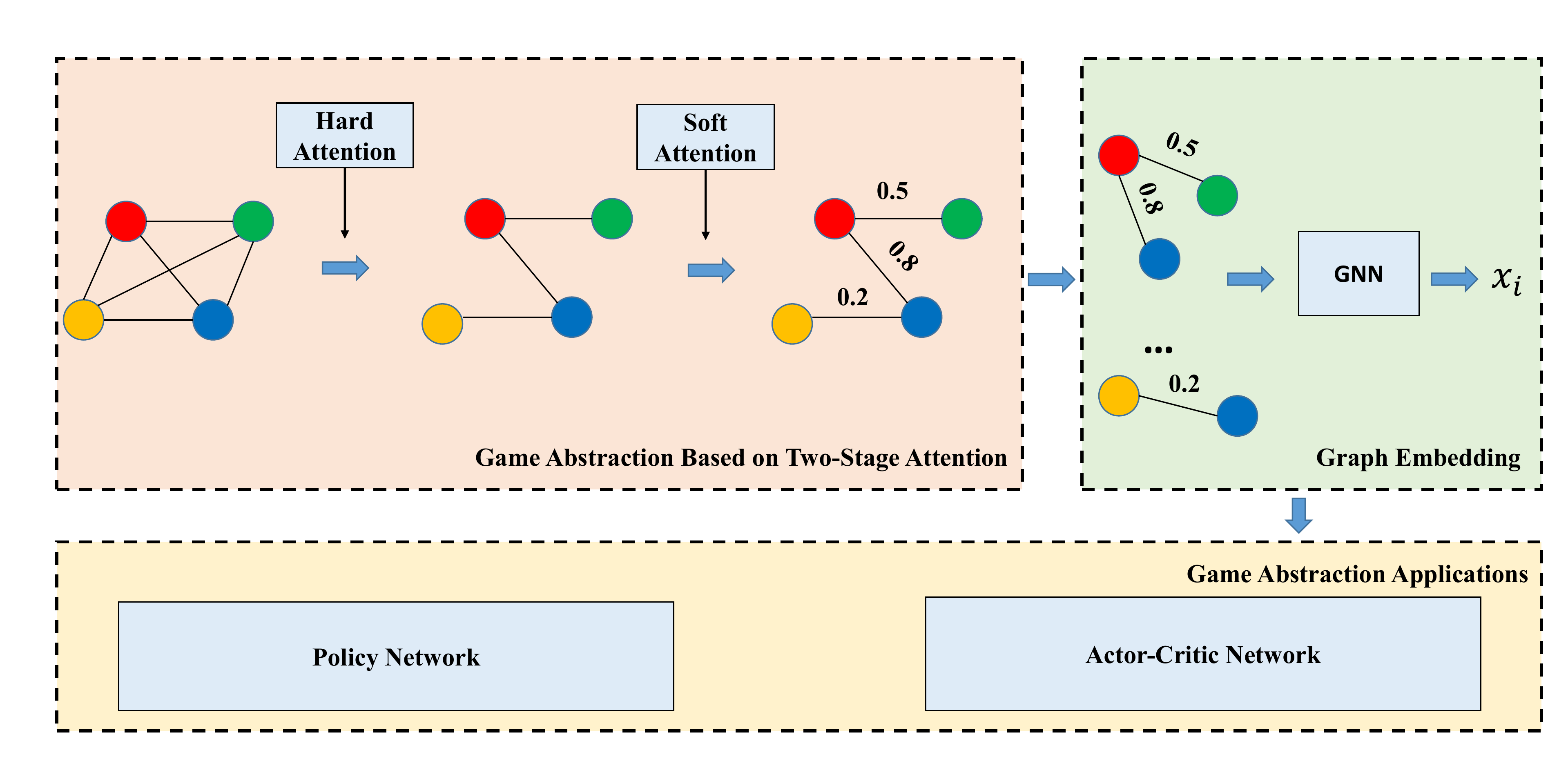}
\caption{Game Abstraction based on two-stage attention mechanism and Graph Neural Network (GNN).}
\label{Fig: idea}
\end{figure}


The key to game abstraction is learning the relationship between agents. Recent work uses soft-attention mechanism to learn the importance distribution of the other agents for each agent \cite{jiang2018learning,iqbal2018actor}. However, the final output softmax function indicates that the importance weight of each agent still depends on the weight of the other agents. That is to say, these methods cannot really learn the relationship between agents and ignore irrelevant agents to simplify policy learning. 


As shown in Figure \ref{Fig: idea}, we represent all agents as a complete graph and propose a novel multi-agent game abstraction algorithm based on two-stage attention network (G2ANet), where hard-attention is used to cut the unrelated edges and soft-attention is used to learn the importance weight of the edges. In addition, we use GNN to obtain the contribution from other agents, which includes the information of the other agents to achieve coordination, and apply the mechanism into several algorithms.
We list the main contributions as follows: 
\begin{itemize}
\item We propose a novel two-stage attention mechanism G2ANet for game abstraction, which can be combined with graph neural network (GNN). 
\item By combining G2ANet with a policy network and $Q$-value network respectively, we propose a communication-based MARL algorithm GA-Comm and an actor-critic (AC) based algorithm GA-AC.
\item Experiments are conducted in Traffic Junction and Predator-Prey. The results show that our methods can simplify the learning process and meanwhile get better asymptotic performance compared with state-of-the-art algorithms.
\end{itemize}

\section{Background}

We review some key concepts in multi-agent reinforcement learning and related work in this section.

\subsection{Markov Game and Game Abstraction}

Markov game, which is also known as stochastic game, is widely adopted as the model of multi-agent reinforcement learning (MARL). It can be treated as the extension of Markov decision process (MDP) to multi-agent setting.

\begin{definition} \label{def: Markov Game}

An n-agent (n $\ge 2$) Markov game is a tuple $\langle N, S, \{A_{i} \}^{n}_{i=1}, \{R_{i}\}^{n}_{i=1}, T \rangle$, where N is the set of agents, S is the state space, $A_{i}$ is the action space of agent i(i=1,...,n). Let $A=A_{1} \times A_{2} \times \cdots \times A_{n}$ be the joint action space. $R_i: S \times A \rightarrow \mathfrak{R}$ is the reward function of agent $i$ and $T: S\times A \times S \rightarrow [0,1]$ is the transition function.

\end{definition}

In a markov game, each agent attempts to maximize its expected sum of discounted rewards, $E\{\sum_{k=0}^{\infty}\gamma^{k}r_{i,t+k} \}$, where $r_{i,t+k}$ is the reward received $k$ steps into the future by agent $i$ and $\gamma$ is the discount factor.

Denote the policy of agent $i$ by $\pi_i: S \times A_i \rightarrow [0,1]$ and the joint policy of all agents by $\pi = (\pi_{1}, \dots , \pi_{n})$. The state-action value function of an agent $i$ under a joint policy $\pi$ can be defined as:

\begin{equation}
Q_{i}^{\pi}(s,\vec a) = E_{\pi} \left \{\sum_{k=0}^{\infty}\gamma^{k}r_{i}^{t+k}|s_{t}=s, \vec a_{t} = \vec a \right \},
\end{equation}

\noindent
where $\vec a \in A$ represents a joint action and $r_{i}^{t+k}$ is the reward received by agent $i$ at time step $(t + k)$. However, since $Q_{i}^{\pi}$ depends on the actions of all agents, the concept of optimal policy should be replaced with joint policy.

\subsubsection{Game Abstraction}

The main idea of game abstraction is to simplify the problem model of multi-agent reinforcement learning (Markov game) to a smaller game, so as to reduce the complexity of solving (or learning) the game equilibrium policy. 


\subsection{Attention Mechanism}


Attention is widely used in many AI fields, including natural language processing \cite{bahdanau2014neural}, computer vision \cite{wang2018non}, and so on. Soft and hard attention are the two major types of attention mechanisms. 

\subsubsection{Soft-Attention} Soft attention calculates a importance distribution of elements.
Specially, soft attention mechanism is fully differentiable and thus can be easily trained by end-to-end back-propagation. Softmax function is a common activation function. However, the function usually assigns nonzero probabilities to unrelated elements, which weakens the attention given to the truly important elements.

\subsubsection{Hard-Attention} Hard attention selects a subset from input elements, which force a model to focus solely on the important elements, entirely discarding the others. 
However, hard attention mechanism is to select elements based on sampling and thus is non-differentiable. Therefore, it cannot learn the attention weight directly through end-to-end back-propagation. 


\subsection{Deep Multi-Agent Reinforcement Learning}

With the development of deep reinforcement learning, recent work in MARL has started moving from tabular methods to deep learning methods. In this paper, we select communication-based algorithms CommNet \cite{sukhbaatar2016learning}, IC3Net \cite{singh2018learning}, Actor-Critic-based algorithms MADDPG \cite{lowe2017multi} and MAAC \cite{iqbal2018actor} as baselines.

\subsubsection{CommNet} CommNet allows communication between agents over a channel where an agent is provided with the average of hidden state representations of the other agents as a communication signal.

\subsubsection{IC3Net} IC3Net can learn when to communicate based on a gating mechanism. The gating mechanism allows agents to block their communication and can be treated as a simple hard-attention.

\subsubsection{MADDPG} 

MADDPG adopts the framework of centralized training with decentralized execution, which allows the policies to use extra information at training time. It is a simple extension of actor-critic policy gradient methods where the critic is augmented with extra information for other agents, while the actor only has access to local information.

\subsubsection{MAAC} MAAC learns a centralized critic with an soft-attention mechanism. The mechanism is able to dynamically select which agents to attend to at each time step.

\section{Our Method}

In this section, we propose a novel game abstraction approach based on two-stage attention mechanism (G2ANet). Based on the mechanism, we propose two novel MARL algorithms (GA-Comm and GA-AC).

\subsection{G2ANet: Game Abstraction Based on Two-Stage Attention}



We construct the relationship between agents as a graph, where each node represents a single agent, and all nodes are connected in pairs by default. We define the graph as Agent-Coordination Graph.

\begin{definition} \label{Def:ACG}
	(Agent-Coordination Graph)
	The relationship between agents is defined as an undirected graph as $G=(N,E)$, consisting of the set $N$ of nodes and the set $E$ of edges, which are unordered pairs of elements of $N$. Each node represents the agent entry, and the edge represents the relationship between the two adjacent agents.
\end{definition}

In large scale multi-agent systems, the number of agents is large, and not all agents need to interact with each other. In this paper, we try to identify unrelated agents by learning the relationship between agents, and perform game abstraction according to the learnt relationship. The simplest way of game abstraction is to design some artificial rules. \citeauthor{yang2018mean} proposed mean-field based multi-agent learning algorithm, where each agent has its own vision and just needs to interact with the agents in its vision \cite{yang2018mean}. However, such mean-field MARL algorithm requires strong prior knowledge of the environment and may not be suitable for application in complex environments. In a large-scale MAS, the interaction between agents is more complicated, the pre-defined rules are difficult to obtain and it cannot dynamically adjust based on the state transition. Inspired by attention mechanism \cite{bahdanau2014neural,ba2014multiple,mnih2014recurrent,xu2015show,vaswani2017attention}, we firstly propose the two-stage attention game abstraction algorithm called G2ANet, which learns the interaction relationship between agents through hard-attention and soft-attention mechanisms.




Recent work has tried to combine MARL with the attention mechanism \cite{jiang2018learning,iqbal2018actor}. However, the main idea is to use the soft-attention mechanism to learn the importance distribution of all other agents to the current agent through softmax function:

\begin{equation}
w_{k}=\frac{exp(f(T,e_{k}))}{\sum_{i=1}^{K}exp(f(T,e_{i}))},
\end{equation}

\noindent
where $e_{k}$ is the feature vector of agent $k$, $T$ is the current agent feature vector, and $w_{k}$ is the importance weight for agent k.


However, the output value of the softmax function is a relative value and cannot really model the relationship between agents. In addition, this method cannot directly reduce the number of agents that need to interact since the unrelated agents will also obtain an importance weight. In addition, the softmax function usually assigns small but nonzero probabilities to trivial agents, which weakens the  attention given to the few truly significant agents. 
In this paper, we propose a novel attention mechanism based on two-stage attention (G2ANet) to solve the above problems.

\begin{figure}[t]
\centering
\includegraphics[width=3.3in]{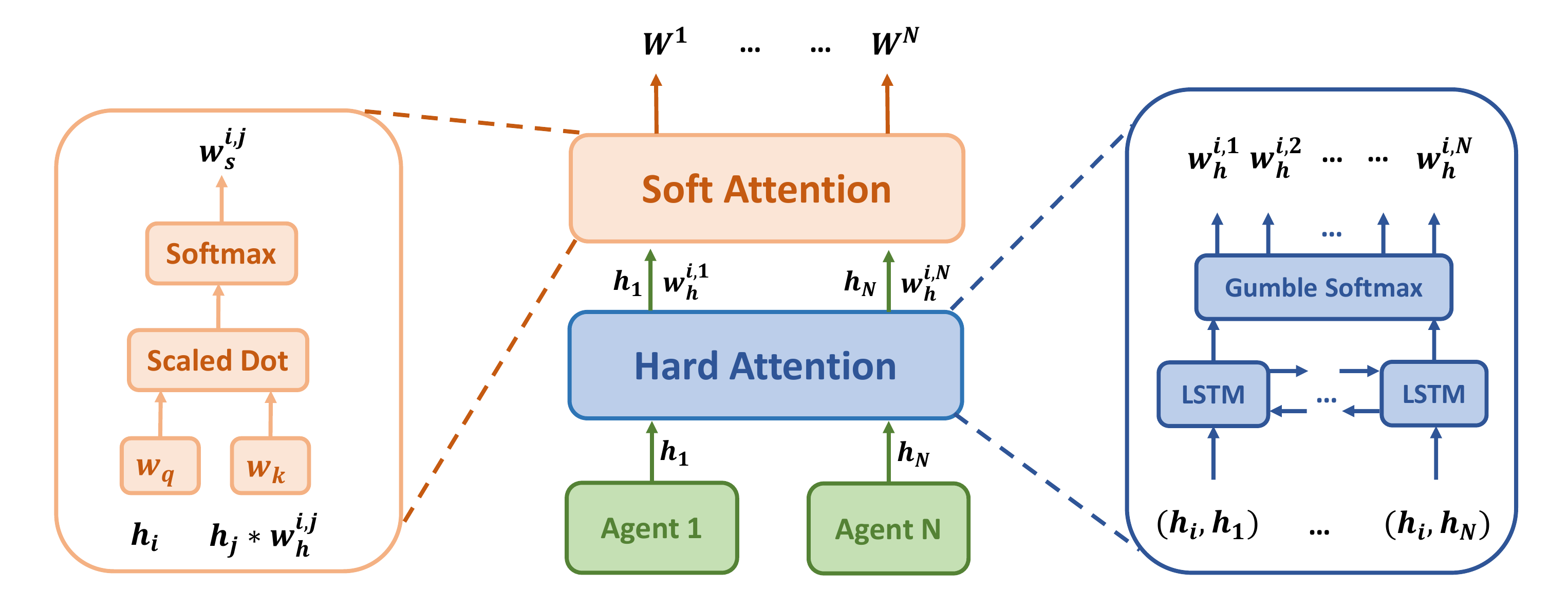}
\caption{Two-Stage Attention Neural Network.}
\label{Fig: Two-Stage-Attention}
\end{figure}

We consider a partially observable environment, where at each time-step $t$, each agent $i$ receives a local observation $o_i^t$, which is the property of agent $i$ in the agent-coordination graph $G$. The local observation $o_i^t$ is encoded into a feature vector $h_i^t$ by MLP. Then, we use the feature vector $h_i^t$ to learn the relationship between agents by attention mechanism. We know that the hard attention model can output a one-hot vector. 
That is, we can get whether the edge between node $i$ and $j$ exist in the graph $G$ and which agents each agent needs to interact with. In this way, the policy learning is simplified to several smaller problems and preliminary game abstraction can be achieved. In addition, we find that each agent plays a different role for a specific agent. That is, the weight of each edge in the graph $G$ is different. Inspired by \cite{vaswani2017attention}, we train a soft-attention model to learn the weight of each edge. In this way, we can get a sub-graph $G_{i}$ for agent $i$, where agent $i$ is only connected with the agents that need to interact with, and the weight on edge describes the importance of the relationship. For sub-graph $G_{i}$, we can use Graph Neural Network (GNN) to obtain a vector representation, which represents the contribution from other agents to agent $i$. 
Moreover, G2ANet has a good generality which can combine with communication-based algorithms \cite{sukhbaatar2016learning,singh2018learning} and AC-based algorithms \cite{lowe2017multi,iqbal2018actor}. We will discuss it in the next subsection.

The two-stage attention mechanism is shown in Figure \ref{Fig: Two-Stage-Attention}. First, we use the hard-attention mechanism to learn the hard weight $W_{h}^{i,j}$, which determines whether there is interaction relationship between agent $i$ and $j$. In this paper, we use a LSTM network to achieve it, where each time-step output a weight (0 or 1) for agent $i,j$, where $j \in \{1,...,n\}$ and $i \neq j$. For agent $i$, we merge the embedding vector of agent $i,j$ into a feature $(h_{i},h_{j})$ and input the feature into LSTM model:

\begin{equation}
h_{i,j} = f(LSTM(h_{i},h_{j})),
\end{equation}

\noindent
where $f(\cdot)$ is a fully connected layer for embedding. However, the output of traditional LSTM network only depends on the input of the current time and the previous time but ignores the input information of the later time. That is to say, the order of the inputs (agents) plays an important role in the process and the output weight value cannot take advantage of all agents' information. We think that is short-sighted and not reasonable. In this paper, we select a Bi-LSTM model to solve it. For example, the relationship weight between agent $i$ and agent $j$ also depends on the information of agent $k$ in the environment, where agent $k \in \{1,...,n\}$ and agent $k$ is not in $\{i,j\}$. 

In addition, the hard-attention is often unable to achieve back-propagation of gradients due to the sampling process. We try to use gumbel-softmax \cite{gumbel} function to solve it:

\begin{equation}
W_{h}^{i,j} = gum(f(LSTM(h_{i},h_{j}))),
\end{equation}

\noindent
where $gum(\cdot)$ represents gumbel-softmax function. By hard-attention mechanism, we can get a sub-graph $G_{i}$ for agent $i$, where agent $i$ just connected with the agents that need to coordinate. Then we use soft-attention to learn the weight of each edge in $G_{i}$. As shown in Figure \ref{Fig: Two-Stage-Attention}, the soft-attention weight $W_{s}^{i,j}$ compares the embedding $e_{j}$ with $e_{i}$, using the query-key system (key-value pair) and passes the matching value between these two embeddings into a softmax function:

\begin{equation}
W_{s}^{i,j} \propto exp(e_{j}^{T}W_{k}^{T}W_{q}e_{i}W_{h}^{i,j}),
\end{equation}

\noindent
where $W_{k}$ transforms $e_{j}$ into a key,  $W_{q}$ transforms $e_{i}$ into a query and $W_{h}^{i,j}$ is the hard-attention value.  Finally, the soft-attention weight value $W_{s}^{i,j}$ is the final weight of the edge, which is defined as $W^{i,j}$.

\subsection{Learning Algorithm Based on Game Abstraction}


Through the two-stage attention model, we can get a reduced graph in which each agent (node) is connected only to the agent (node) that needs to interact with. For example, in Figure \ref{Fig: idea}, we can get a sub-graph $G_i$ for agent $i$, where the center node is agent $i$ (node $i$). As we all know, GNN has powerful encoding ability. If each node represents the agent's encoding in the sub-graph $G_i$, we can use GNN to get a joint encoding for agent $i$, which defines the contribution of all other agents for the current agent $i$. By the joint vector encoding, our method can make better decisions.
As mentioned earlier, our two-stage attention-based game abstraction method is a general mechanism.
In this paper, we combine G2ANet with policy network and $Q$-value network respectively, and propose two learning algorithms:
(1) \textbf{Policy network in communication model (GA-Comm)}: Each agent considers the communication vectors of all other agents when making decisions; (2) \textbf{Critic network in actor-critic model (GA-AC)}: Critic network of each agent considers the state and action information of all other agents when calculating its $Q$-value in AC-based methods.

\begin{figure}[tbp]
\centering
\includegraphics[width=3.5in]{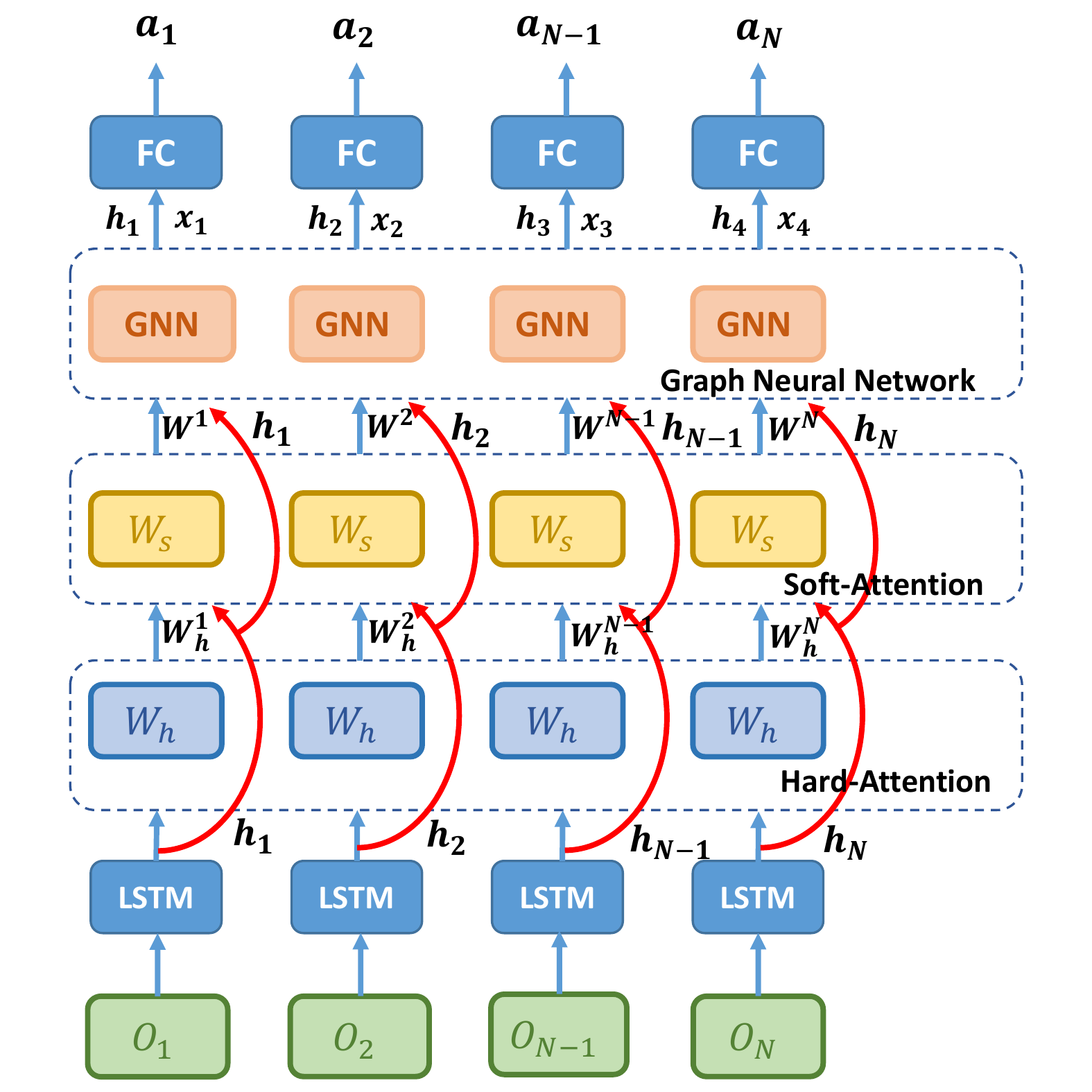}
\caption{Communication model based on Game Abstraction}
\label{Fig: GA-Comm}
\end{figure}

\subsubsection{Policy Network Based on Game Abstraction} \label{GA-Comm}


As we all know, much related work focus on learning multi-agent communication \cite{sukhbaatar2016learning,singh2018learning}, most of which achieve communication through aggregation function, which can access all other agents' communication vector (e.g., average function, maximum function) into one vector and pass it to each agent. In this way, each agent can receive all agent's information and achieve communication. However, there is no need for each agent to communicate with all other agents in most environments. That means the frequent communication will cause high computing cost and increase the difficulty of policy learning. 
In this paper, we combine the novel game abstraction mechanism G2ANet with policy network and propose a novel communication-based MARL learning algorithm GA-Comm.

%
As shown in Figure \ref{Fig: GA-Comm}, $o_{i}$ represents the observation of agent $i$,
its policy takes the form of:

\begin{equation}
a_{i} = \pi(h_{i},x_{i}),
\end{equation}

\noindent
where $\pi$ is the action policy of an agent, $h_{i}$ is the observation feature for agent $i$ and $x_{i}$ is the contribution from other agents for agent $i$. In this paper, we use a LSTM layer to extract the feature:


\begin{equation}
h_{i},s_{i} = LSTM(e(o_{i}),h_{i}^{'},s_{i}^{'}),
\end{equation}

\noindent
where $o_{i}$ is the observation of agent $i$ at time-step $t$, $e(\cdot)$ is an encoder function parameterized by a fully-connected neural network. Also, $h_{i}$ and $s_{i}$ are the hidden and cell states of the LSTM. As for the contribution for agent $i$ from other agents, we firstly use two-stage attention mechanism to select which agents the agent $i$ need to communicate and obtain its importance:

\begin{equation}
W_{h}^{i,j} = M_{hard}(h_{i},h_{j}), \\
W_{s}^{i,j} = M_{soft}(W_{h},h_{i},h_{j}),
\end{equation}

\noindent
where $W_{h}^{i,j}$ is the hard-attention value and $W_{s}^{i,j}$ is the soft-attention value calculated by hidden state $h_{i}$, $h_{j}$. $M_{hard}$ is the hard-attention model and $M_{soft}$ is the soft-attention model. In this way, we can get the contribution $x_{i}$ from other agents by GNN. We use a simple method to calculate, which is a weighted sum of other agents' contribution by two-stage attention mechanism:

\begin{equation}
x_{i} = \sum_{j \ne i}w_{j}h_{j} = \sum_{j \ne i}W_{h}^{i,j}W_{s}^{i,j}h_{j}.
\end{equation}

Finally, we can get the action $a_{i}$ for agent $i$. During the training process, we train the policy $\pi$ with REINFORCE \cite{williams1992simple}.

\subsubsection{Actor-Critic Network Based on Game Abstraction} \label{GA-AC}

Inspired by MAAC \cite{iqbal2018actor}, we propose a novel learning algorithm based on G2ANet. To calculate the $Q$-value $Q_i(o_{i},a_{i})$ for agent $i$, the critic network receives the observations $o=(o_{1},...,o_{N})$ and actions, $a=(a_{1},...,a_{N})$ for all agents. $Q_{i}(o_{i},a_{i})$ is the value function for agent $i$ :

\begin{equation}
Q_{i}(o_{i},a_{i}) = f_{i}(g_{i}(o_{i},a_{i}),x_{i}),
\end{equation}

\noindent
where $f_{i}$ and $g_{i}$ is a multi-layer perception (MLP), $x_{i}$ is the contribution from other agents, which is computed by GNN. In this paper, we use a simple method, which is a weighted sum of each agent’s value based on our two-stage attention mechanism:

\begin{equation}
x_{i} = \sum_{j \ne i}w_{j}v_{j} = \sum_{j \ne i}w_{j}h(Vg_{j}(o_{j},a_{j})),
\end{equation}

\noindent
where the value, $v_j$ is an embedding of agent $j$, encoded with an embedding function and then transformed by a shared matrix $V$ and $h(\cdot)$ is an elementwise nonlinearity.

The attention weight $w_{j}$ is computed by the two-stage attention mechanism, which compares the embedding $e_{j}$ with
$e_{i} = g_{i}(o_{i},a_{i})$ and passes the relation value between these two embeddings into a softmax function:

\begin{equation}
w_{j} = W_{h}^{i,j}W_{s}^{i,j} \propto exp(h(BiLSTM_{j}(e_{i},e_{j}))e_{j}^{T}W_{k}^{T}W_{q}e_{i}),
\end{equation}

\noindent
where $W_q$ transforms $e_i$ into a query and $W_k$ transforms
$e_j$ into a key. In this way, we can obtain the attention weight $w_{j}$ and calculate the $Q$ value for each agent.

\begin{figure}[tbp]
\centering
\includegraphics[width=3.7in]{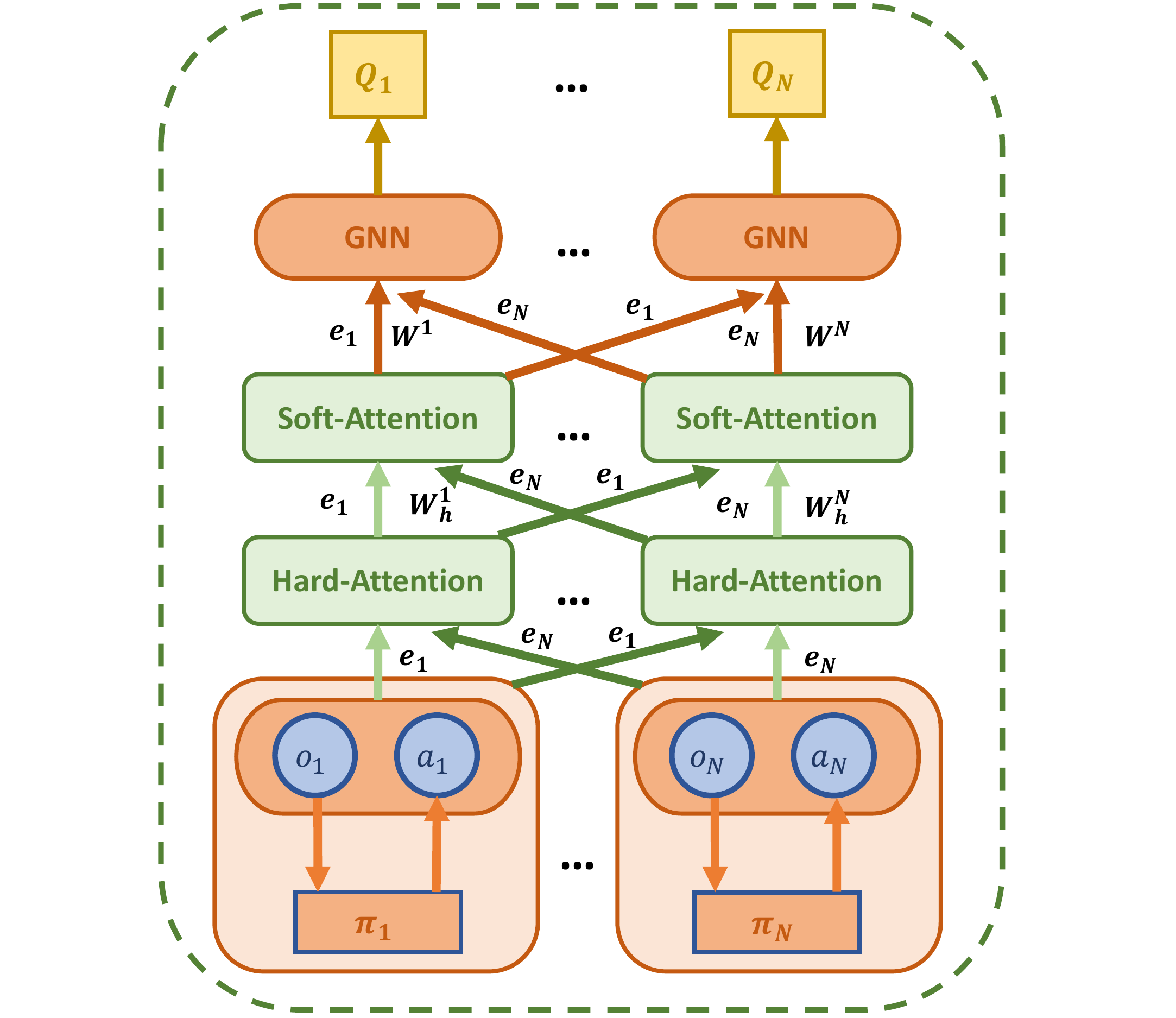}
\caption{Actor-Critic model based on Game Abstraction}
\label{Fig: GA-AC}
\end{figure}

\begin{figure*}[htbp]
\centering
\includegraphics[width=7in]{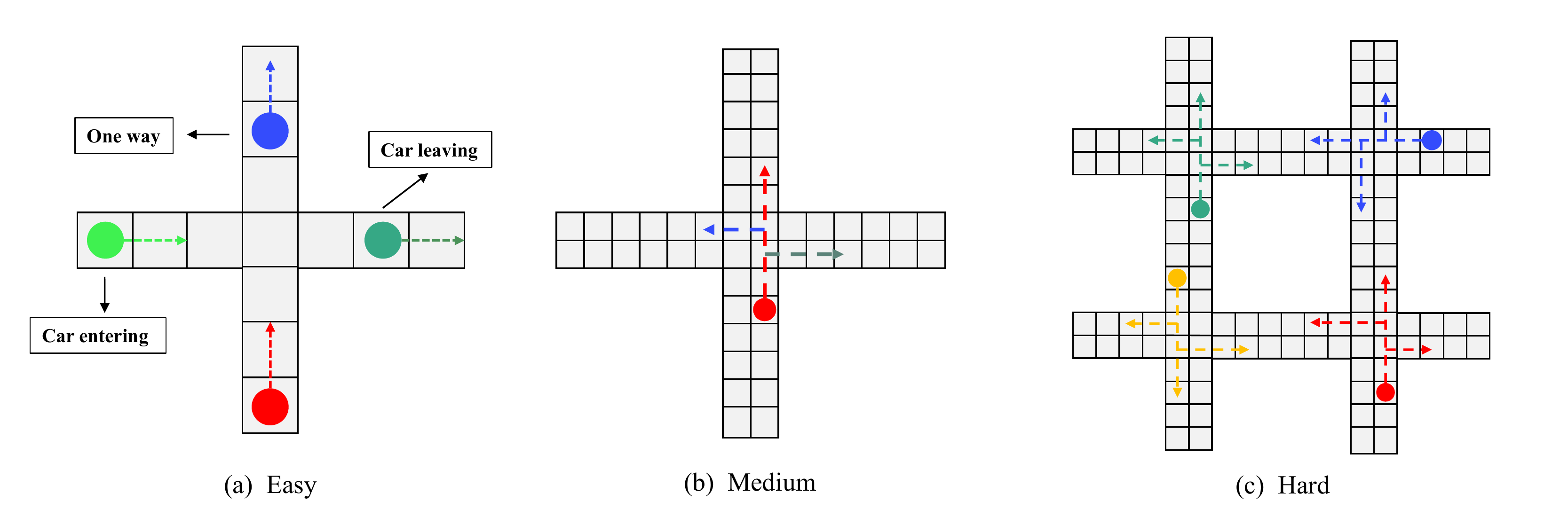}
\caption{Traffic Junction Environment. Agents have zero vision and can only observe their own location. The cars have to cross the the whole road minimizing collisions.}
\label{Fig: Exp_TJ}
\end{figure*}

\section{Experiments}


In this section, we evaluate the performance of our game abstraction algorithms in two scenarios.
The first one is conducted in Traffic Junction \cite{singh2018learning}, where we use policy based game abstraction algorithm GA-Comm and baselines are CommNet and IC3Net. The second is the Predator-Prey in Multi-Agent Particle environment \cite{lowe2017multi}, where we use $Q$-value based game abstraction algorithm GA-AC and the baselines are MADDPG and MAAC. 



\subsection{Traffic Junction}
The simulated traffic junction environments from \cite{singh2018learning} consists of cars moving along pre-assigned potentially interesting routes on one or more road junctions. “Success” indicates that no collisions occur at a time-step. We can calculate the success rate according to the number of time steps and collisions (failures) in each episode.
The total number of cars is fixed at $N_{max}$ and new cars get added to the environment with probability $p_{arrive}$ at every time-step.
The task has three difficulty levels which vary in the number of possible routes, entry points and junctions. Fallowing the same setting in IC3Net \cite{singh2018learning}, the number of agents in the easy, medium, and hard environments is 5, 10, and 20, respectively. We make this task harder by always setting vision to zero in all the three difficulty levels, which means that each agent's local observation only has its position information and each agent need to obtain other agents’ information to achieve coordination via communication mechanism.
The action space for each car is gas and break, and the reward consists of a linear time penalty $-0.01\tau$, where $\tau$ is the number of time-steps since the car has been active, and a collision penalty $r_{collision} = -10$.

\begin{figure*}[htbp]
\centering
\includegraphics[width=7in]{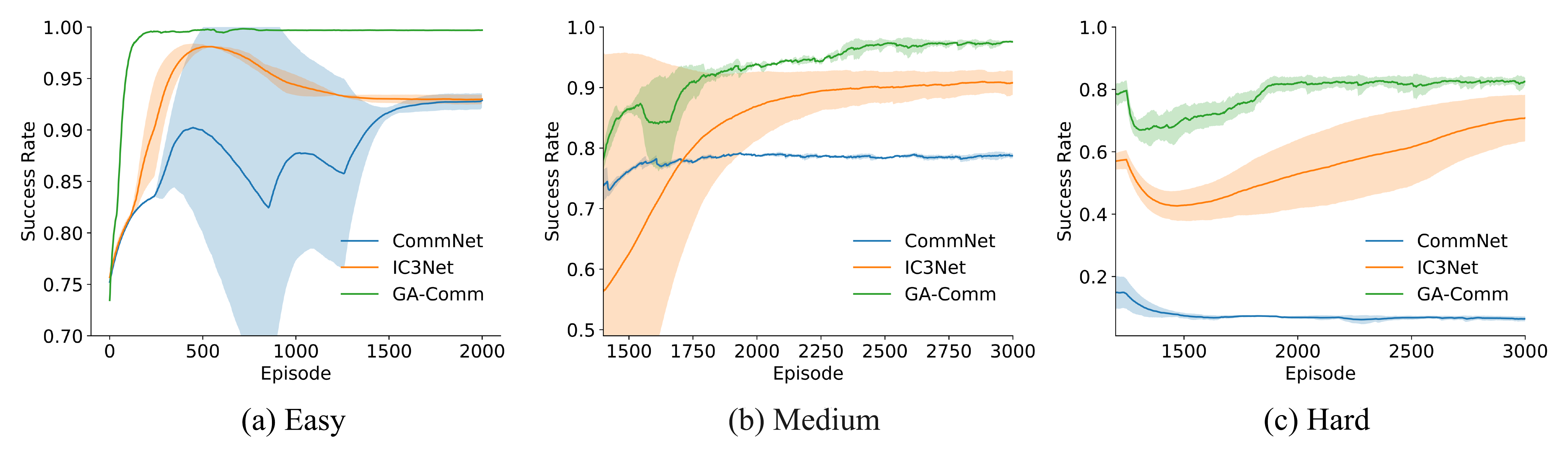}
\caption{Experimental results in Traffic Junction. (a) is the result in easy version, (b) is the result in medium version and (c) is the result in hard version. Shaded regions are one standard deviation over 10 runs.}
\label{Fig: Exp_TJ_Result}
\end{figure*}

Figure \ref{Fig: Exp_TJ_Result} illustrates the success rate per episode attained by various methods on TJ, where GA-Comm is our communication model based on G2ANet and IC3Net is a communication method based on one-stage hard attention. Table 1 shows the success rates on different levels (easy, medium, and hard), which is the average success rate of 10 runs and the variance of the 10 repeated experiments can be obtained from the shaded area in Figure \ref{Fig: Exp_TJ_Result}.
Our proposed approach based on game abstraction is competitive when compared to other methods.


As the setting in IC3Net \cite{singh2018learning}, we use the method of curriculum learning to train the model, gradually increase the number of agents in the environment, and further simplify the learning of the model. As shown in Figure \ref{Fig: Exp_TJ_Result}, GA-Comm performs better than all baseline methods in all modes. Our approach is not only high in success rate but also more stable. In addition, as the difficulty of the environment gradually increases (the number of junctions increases) and the number of agents gradually increases, the effect is more obvious. We can find that the success rate of our method is about 6\%, 7\% and 11\% higher than IC3Net in the three levels (easy, medium and hard), which verifies that our method is more effective (6-7-11) as the difficulty of environment gradually increases. This further illustrates the applicability of our game abstraction mechanism in large-scale multi-agent systems.

\begin{figure}[tbp]
\centering
\includegraphics[width=3.7in]{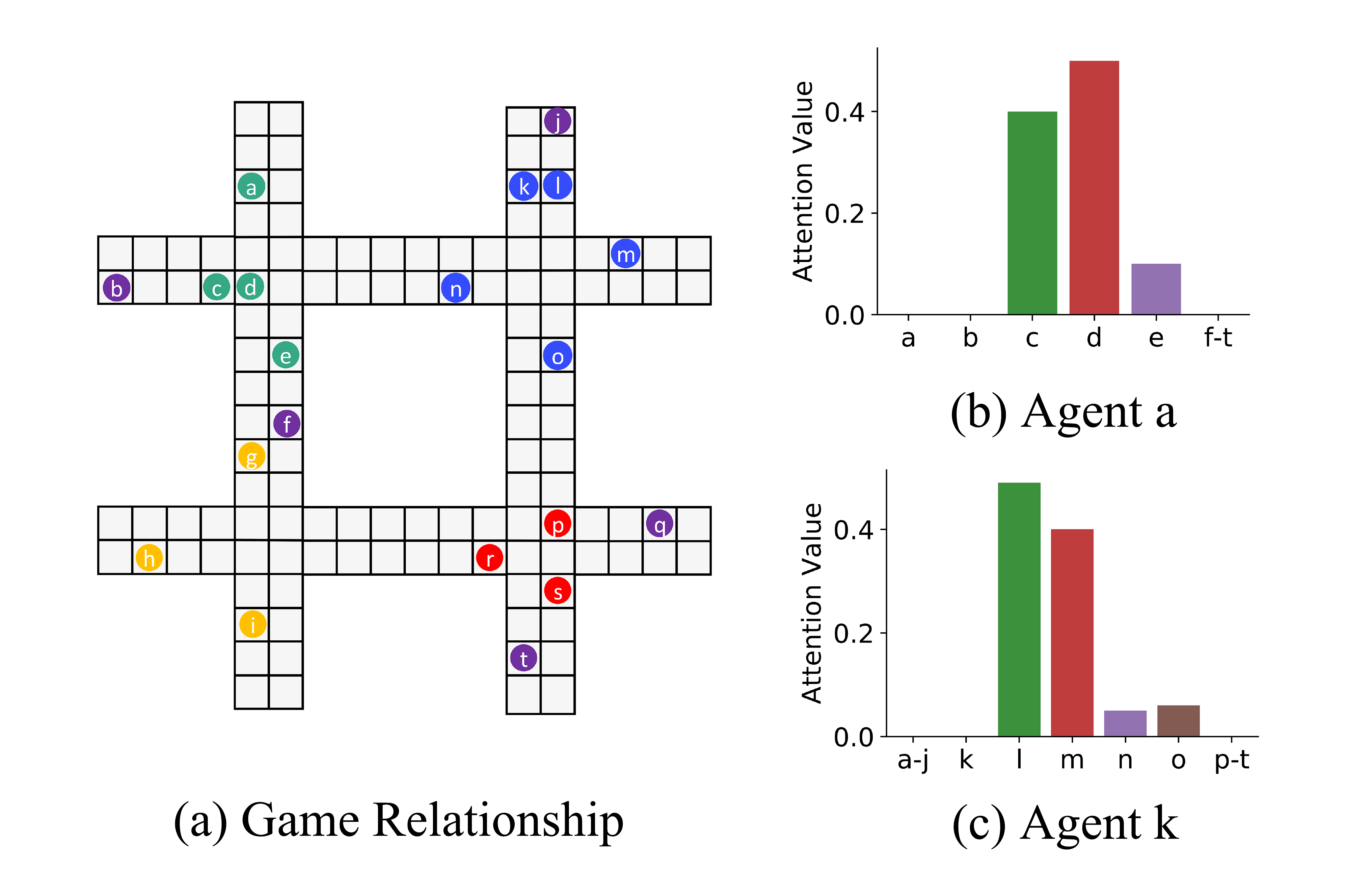}
\caption{Agents with the same color represent a group, and each agent just need to interact with the agents in the group.}
\label{Fig: Exp_TJ_Display}
\end{figure}

At different time steps in an episode, the relationship between agents is constantly changing. In our method, we can learn the adaptive and dynamic attention value. In order to analyze the influence of the game abstraction mechanism on the learning results, the game relationship between agents is showed in Figure \ref{Fig: Exp_TJ_Display}(a), which only describes the attention values of one certain time-step. Each agent has its color (e.g., green, blue, yellow, red, and purple), and the same color agents represent a group. It is observed that each agent can select their partners and form a group (purple is the independent agent), and ignores the unrelated agents. For example, all agents are mainly divided into four groups, each group mainly gathers near the junction. For agent $a$, the green agents are its teammates, which concentrate on one junction, and it can ignore other agents when making a decision. In addition, for each agent, the importance is different for the agents in the group. Figure \ref{Fig: Exp_TJ_Display}(b-c) shows the final attention value distribution for agent $a$ (left) and agent $k$. Agent $a,c,d,e$ in the same group and the importance of agent $c$ and agent $d$ is larger than agent $e$ for agent $a$. Similarly, the importance of agent $l,m$ is larger than agent $n,o$ for agent $k$. 
We can conclude that the game abstraction that first ignores unrelated agents, and then learns an important distribution in a smaller number of environments.
In this way, we can avoid learning the importance distribution of all agents directly in a larger-scale MAS, and the final value is more accurate.

\begin{table}[]
    \caption{Success Rate in the Traffic Junction}

    \centering
    \begin{tabular}{p{2cm}p{1.5cm}p{1.5cm}p{1.5cm}p{1.25cm}}
    \toprule[1pt]
        Algorithm & Easy & Medium & Hard \\
        \hline
        CommNet  &93.5\%    & 78.8\%             & 6.5\%                      \\
        IC3Net   &93.2\%    & 90.8\%             & 70.9\%                         \\
        GA-Comm  &\textbf{99.7\%}    & \textbf{97.6\%}          & \textbf{82.3\%}                       \\
    \bottomrule[1pt]    
    \end{tabular}
    \label{Tab: Exp_TJ}
\end{table}

\subsection{Multi-Agent Particle Environment}

\begin{figure}[htbp]
\centering
\includegraphics[width=\linewidth]{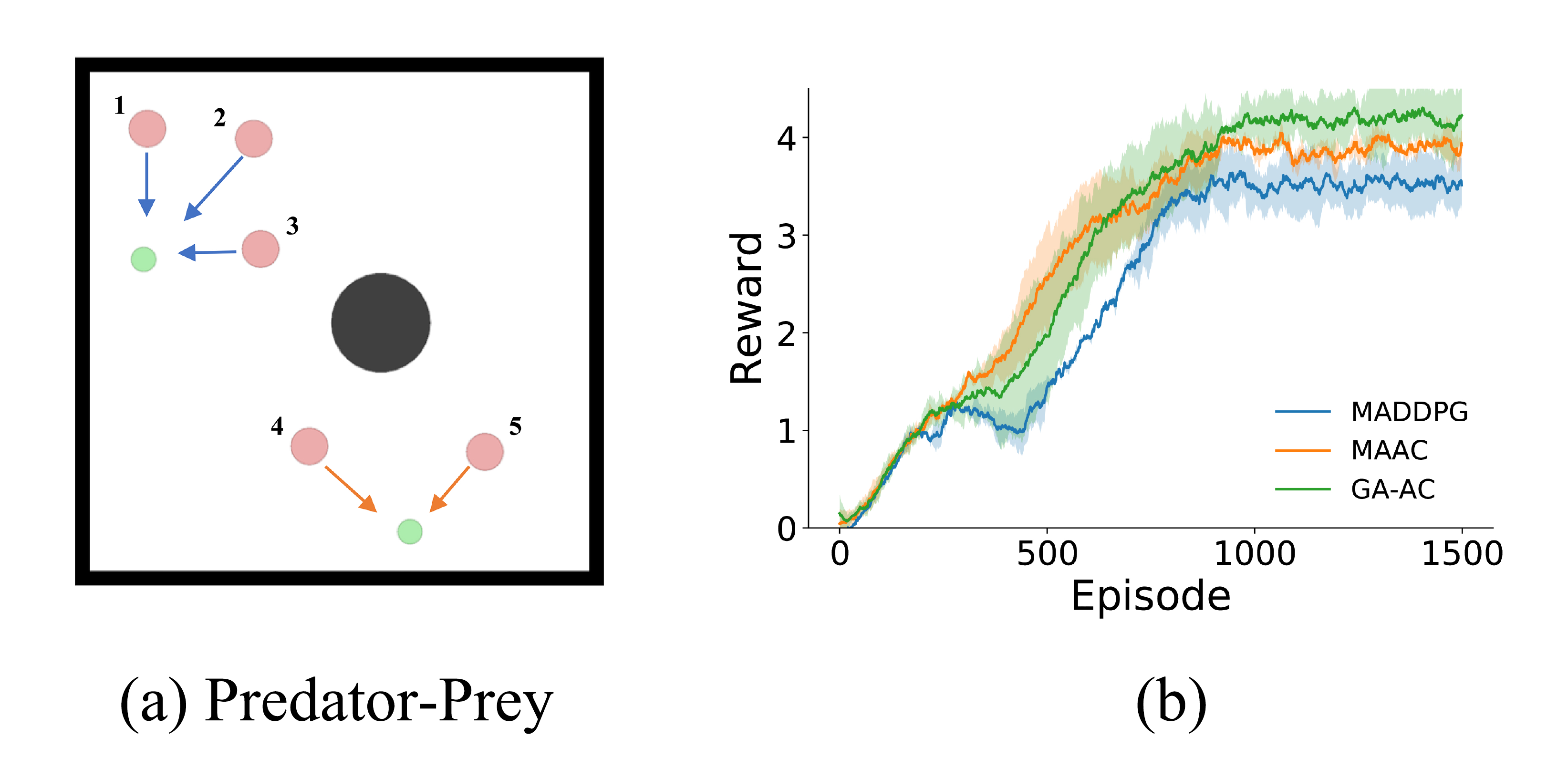}
\caption{Experimental result in Predator-Prey.}
\label{Fig: Exp_PP_Result}
\end{figure}

The second scenario in this paper is the Multi-Agent Particle Environment. 
As shown in Figure \ref{Fig: Exp_PP_Result}(a), we choose $predator-prey$ as the test environment, where the adversary agent (red) is slower and needs to capture the good agent (green), and the good agent is faster and needs to escape. In this paper, we fix the policy (DQN) of the good agents. As the setting in MADDPG, adversary agents receive a reward of +10 when they capture good agents.

 We trained the model in the setting of $N_{a} = 5$ and $N_{g} = 2$ for 1500 episodes, where $N_{a}$ is the number of adversary and $N_{g}$ is the number of good agents. Similarly, adversary agents need to achieve multiple groups to capture all the good agents. Figure \ref{Fig: Exp_PP_Result} shows the learning curves of each agent's average reward, where MADDPG is an algorithm proposed by \citeauthor{lowe2017multi} and MAAC is a soft-attention based algorithm proposed by \citeauthor{iqbal2018actor}. GA-AC outperforms all the baselines in terms of mean reward. It is observed that our method is slower to learn (Compared with the soft-attention method MAAC) in the early stage. We think that is because the architecture of our two-stage attention network is more complex. The final better performance verifies the effectiveness of our game abstraction mechanism.

\begin{figure}[!htp]
\centering
\includegraphics[width=\linewidth]{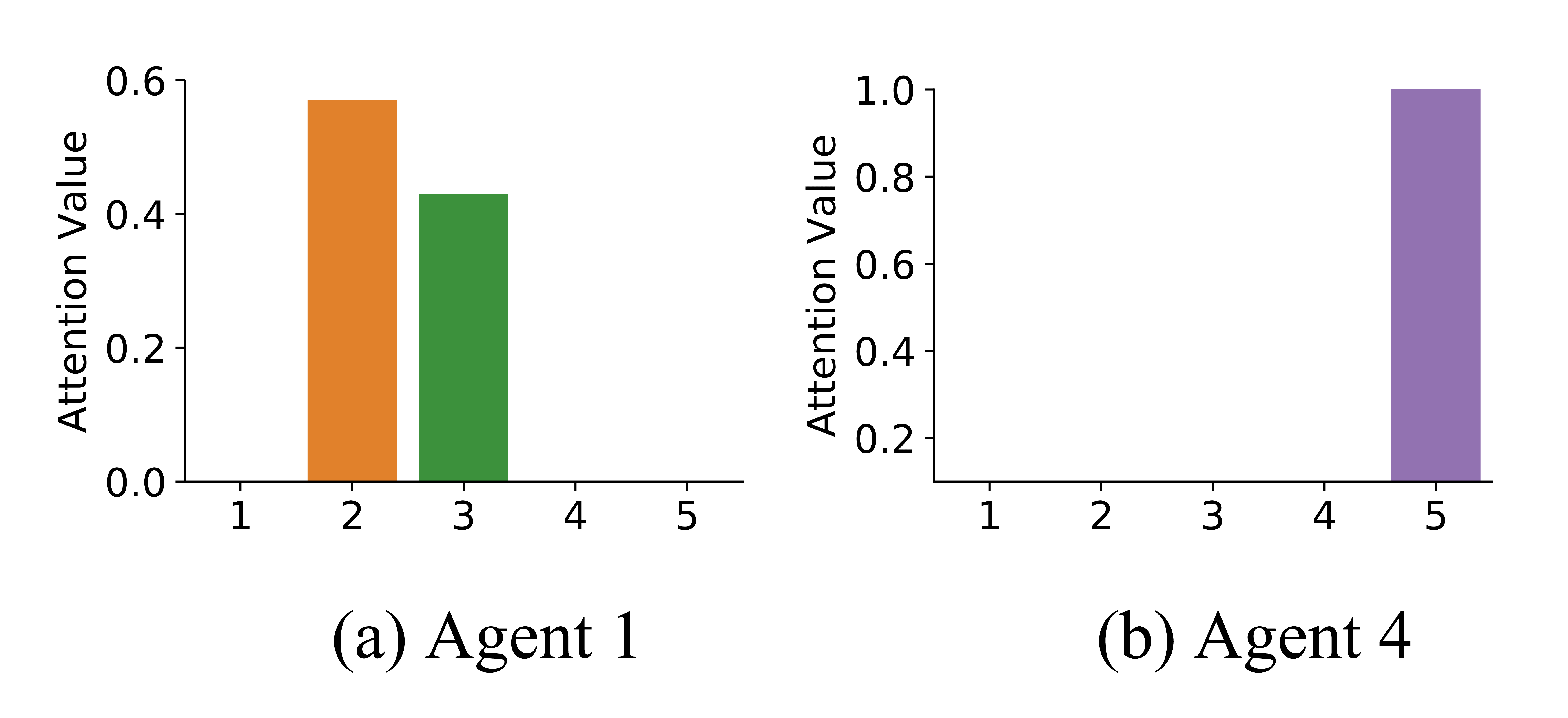}
\caption{Attention value distribution. (a) is the attention contribution for agent 1, (b) is the attention distribution for agent 4.}
\label{Fig: PP_Attention}
\end{figure}

As shown in Figure \ref{Fig: Exp_PP_Result}, it's observed that five adversary agents are divided into two groups to chase two good agents. Each agent just needs to interact with the agents in the same group, which can effectively avoid the interference of the unrelated agents. The final result also shows that our game abstraction mechanism based algorithm GA-AC has learned a reasonable combination form. In Figure \ref{Fig: PP_Attention}, we can obtain the attention value distribution for agent $1$ (Figure \ref{Fig: PP_Attention}(a)) and agent $4$ (Figure \ref{Fig: PP_Attention}(b)). Agents $1,2,3$ are in the same group, and the importance of agent $2$ and agent $3$ is larger than that of agents 4 and 5 for agent $1$. Similarly, the importance of agent $5$ is larger than agent that of agents 1, 2, and 3 for agent $4$. We can conclude that the game abstraction method proposed in this paper can well model the game relationship between agents, avoid the interference of unrelated agents and accelerate the process of policy learning.

\section{Conclusions}
In this paper, we focus on the simplification of policy learning in large-scale multi-agent systems. We learn the relationship between agents and achieve game abstraction by defining a novel attention mechanism. 
At different time steps in an episode, the relationship between agents is constantly changing. In this paper, we can learn the adaptive and dynamic attention value.
Our major contributions include the novel two-stage attention mechanism G2ANet, and the two game abstraction based learning algorithms GA-Comm and GA-AC. Experimental results in Traffic Junction and Predator-Prey show that with the novel game abstraction mechanism, the GA-Comm and GA-AC algorithms can get better performance compared with state-of-the-art algorithms. 

\section{Acknowledgments}

This work is supported by Science and Technology Innovation 2030 –“New Generation Artificial Intelligence” Major Project No.(2018AAA0100905), the National Natural Science Foundation of China (Nos.: 61432008, 61702362, U1836214, 61403208), the Collaborative Innovation Center of Novel Software Technology and Industrialization.

\bibliographystyle{aaai}
\bibliography{reference}

\begin{thebibliography}{}

\bibitem[\protect\citeauthoryear{Ba, Mnih, and
  Kavukcuoglu}{2014}]{ba2014multiple}
Ba, J.; Mnih, V.; and Kavukcuoglu, K.
\newblock 2014.
\newblock Multiple object recognition with visual attention.
\newblock {\em arXiv preprint arXiv:1412.7755}.

\bibitem[\protect\citeauthoryear{Bahdanau, Cho, and
  Bengio}{2014}]{bahdanau2014neural}
Bahdanau, D.; Cho, K.; and Bengio, Y.
\newblock 2014.
\newblock Neural machine translation by jointly learning to align and
  translate.
\newblock {\em arXiv preprint arXiv:1409.0473}.

\bibitem[\protect\citeauthoryear{Chen \bgroup et al\mbox.\egroup
  }{2018}]{chen2018factorized}
Chen, Y.; Zhou, M.; Wen, Y.; Yang, Y.; Su, Y.; Zhang, W.; Zhang, D.; Wang, J.;
  and Liu, H.
\newblock 2018.
\newblock Factorized q-learning for large-scale multi-agent systems.
\newblock {\em arXiv preprint arXiv:1809.03738}.

\bibitem[\protect\citeauthoryear{De~Hauwere, Vrancx, and
  Now{\'e}}{2010}]{de2010learning}
De~Hauwere, Y.-M.; Vrancx, P.; and Now{\'e}, A.
\newblock 2010.
\newblock Learning multi-agent state space representations.
\newblock In {\em Proceedings of the 9th International Conference on Autonomous
  Agents and Multiagent Systems},  715--722.

\bibitem[\protect\citeauthoryear{Foerster \bgroup et al\mbox.\egroup
  }{2018}]{foerster2018counterfactual}
Foerster, J.~N.; Farquhar, G.; Afouras, T.; Nardelli, N.; and Whiteson, S.
\newblock 2018.
\newblock Counterfactual multi-agent policy gradients.
\newblock In {\em Thirty-Second AAAI Conference on Artificial Intelligence}.

\bibitem[\protect\citeauthoryear{Guestrin, Lagoudakis, and
  Parr}{2002}]{GuestrinLP02}
Guestrin, C.; Lagoudakis, M.~G.; and Parr, R.
\newblock 2002.
\newblock Coordinated reinforcement learning.
\newblock In {\em Proceedings of the 9th International Conference on Machine
  Learning},  227--234.

\bibitem[\protect\citeauthoryear{Hu, Gao, and An}{2015}]{hu2015learning}
Hu, Y.; Gao, Y.; and An, B.
\newblock 2015.
\newblock Learning in multi-agent systems with sparse interactions by knowledge
  transfer and game abstraction.
\newblock In {\em {Proceedings of the 2015 International Conference on
  Autonomous Agents and Multiagent Systems}},  753--761.

\bibitem[\protect\citeauthoryear{Iqbal and Sha}{2019}]{iqbal2018actor}
Iqbal, S., and Sha, F.
\newblock 2019.
\newblock Actor-attention-critic for multi-agent reinforcement learning.
\newblock In {\em Proceedings of the 36th International Conference on Machine
  Learning},  2961--2970.

\bibitem[\protect\citeauthoryear{Jang, Gu, and Poole}{2017}]{gumbel}
Jang, E.; Gu, S.; and Poole, B.
\newblock 2017.
\newblock Categorical reparameterization with gumbel-softmax.
\newblock In {\em 5th International Conference on Learning Representations}.

\bibitem[\protect\citeauthoryear{Jiang and Lu}{2018}]{jiang2018learning}
Jiang, J., and Lu, Z.
\newblock 2018.
\newblock Learning attentional communication for multi-agent cooperation.
\newblock In {\em Advances in Neural Information Processing Systems},
  7254--7264.

\bibitem[\protect\citeauthoryear{Jiang, Dun, and Lu}{2018}]{jiang2018graph}
Jiang, J.; Dun, C.; and Lu, Z.
\newblock 2018.
\newblock Graph convolutional reinforcement learning for multi-agent
  cooperation.
\newblock {\em arXiv preprint arXiv:1810.09202}.

\bibitem[\protect\citeauthoryear{Kok and Vlassis}{2004}]{KokV04}
Kok, J.~R., and Vlassis, N.~A.
\newblock 2004.
\newblock Sparse cooperative {Q}-learning.
\newblock In {\em Proceedings of the 21st International Conference on Machine
  Learning},  61--68.

\bibitem[\protect\citeauthoryear{Liu \bgroup et al\mbox.\egroup
  }{2019}]{yong2019ijcai}
Liu, Y.; Hu, Y.; Gao, Y.; Chen, Y.; and Fan, C.
\newblock 2019.
\newblock Value function transfer for deep multi-agent reinforcement learning
  based on n-step returns.
\newblock In {\em Proceedings of the Twenty-Eighth International Joint
  Conference on Artificial Intelligence},  457--463.

\bibitem[\protect\citeauthoryear{Lowe \bgroup et al\mbox.\egroup
  }{2017}]{lowe2017multi}
Lowe, R.; Wu, Y.; Tamar, A.; Harb, J.; Abbeel, O.~P.; and Mordatch, I.
\newblock 2017.
\newblock Multi-agent actor-critic for mixed cooperative-competitive
  environments.
\newblock In {\em Advances in Neural Information Processing Systems},
  6379--6390.

\bibitem[\protect\citeauthoryear{Melo and Veloso}{2011}]{MeloV11}
Melo, F.~S., and Veloso, M.~M.
\newblock 2011.
\newblock Decentralized {MDPs} with sparse interactions.
\newblock {\em Artifitial Intelligence} 175(11):1757--1789.

\bibitem[\protect\citeauthoryear{Mnih \bgroup et al\mbox.\egroup
  }{2014}]{mnih2014recurrent}
Mnih, V.; Heess, N.; Graves, A.; et~al.
\newblock 2014.
\newblock Recurrent models of visual attention.
\newblock In {\em Advances in neural information processing systems},
  2204--2212.

\bibitem[\protect\citeauthoryear{Mnih \bgroup et al\mbox.\egroup
  }{2016}]{mnih2016asynchronous}
Mnih, V.; Badia, A.~P.; Mirza, M.; Graves, A.; Lillicrap, T.; Harley, T.;
  Silver, D.; and Kavukcuoglu, K.
\newblock 2016.
\newblock Asynchronous methods for deep reinforcement learning.
\newblock In {\em International conference on machine learning},  1928--1937.

\bibitem[\protect\citeauthoryear{Rashid \bgroup et al\mbox.\egroup
  }{2018}]{rashid2018qmix}
Rashid, T.; Samvelyan, M.; de~Witt, C.~S.; Farquhar, G.; Foerster, J.~N.; and
  Whiteson, S.
\newblock 2018.
\newblock {QMIX:} monotonic value function factorisation for deep multi-agent
  reinforcement learning.
\newblock In {\em Proceedings of the 35th International Conference on Machine
  Learning},  4292--4301.

\bibitem[\protect\citeauthoryear{Schulman \bgroup et al\mbox.\egroup
  }{2017}]{schulman2017proximal}
Schulman, J.; Wolski, F.; Dhariwal, P.; Radford, A.; and Klimov, O.
\newblock 2017.
\newblock Proximal policy optimization algorithms.
\newblock {\em arXiv preprint arXiv:1707.06347}.

\bibitem[\protect\citeauthoryear{Singh, Jain, and
  Sukhbaatar}{2019}]{singh2018learning}
Singh, A.; Jain, T.; and Sukhbaatar, S.
\newblock 2019.
\newblock Learning when to communicate at scale in multiagent cooperative and
  competitive tasks.
\newblock In {\em 7th International Conference on Learning Representations}.

\bibitem[\protect\citeauthoryear{Sukhbaatar, Fergus, and
  others}{2016}]{sukhbaatar2016learning}
Sukhbaatar, S.; Fergus, R.; et~al.
\newblock 2016.
\newblock Learning multiagent communication with backpropagation.
\newblock In {\em Advances in Neural Information Processing Systems},
  2244--2252.

\bibitem[\protect\citeauthoryear{Sunehag \bgroup et al\mbox.\egroup
  }{2018}]{sunehag2018value}
Sunehag, P.; Lever, G.; Gruslys, A.; Czarnecki, W.~M.; Zambaldi, V.; Jaderberg,
  M.; Lanctot, M.; Sonnerat, N.; Leibo, J.~Z.; Tuyls, K.; et~al.
\newblock 2018.
\newblock Value-decomposition networks for cooperative multi-agent learning
  based on team reward.
\newblock In {\em Proceedings of the 17th International Conference on
  Autonomous Agents and MultiAgent Systems},  2085--2087.

\bibitem[\protect\citeauthoryear{Vaswani \bgroup et al\mbox.\egroup
  }{2017}]{vaswani2017attention}
Vaswani, A.; Shazeer, N.; Parmar, N.; Uszkoreit, J.; Jones, L.; Gomez, A.~N.;
  Kaiser, {\L}.; and Polosukhin, I.
\newblock 2017.
\newblock Attention is all you need.
\newblock In {\em Advances in neural information processing systems},
  5998--6008.

\bibitem[\protect\citeauthoryear{Wang \bgroup et al\mbox.\egroup
  }{2018}]{wang2018non}
Wang, X.; Girshick, R.; Gupta, A.; and He, K.
\newblock 2018.
\newblock Non-local neural networks.
\newblock In {\em Proceedings of the IEEE Conference on Computer Vision and
  Pattern Recognition},  7794--7803.

\bibitem[\protect\citeauthoryear{Williams}{1992}]{williams1992simple}
Williams, R.~J.
\newblock 1992.
\newblock Simple statistical gradient-following algorithms for connectionist
  reinforcement learning.
\newblock {\em Machine learning} 8(3-4):229--256.

\bibitem[\protect\citeauthoryear{Xu \bgroup et al\mbox.\egroup
  }{2015}]{xu2015show}
Xu, K.; Ba, J.; Kiros, R.; Cho, K.; Courville, A.; Salakhudinov, R.; Zemel, R.;
  and Bengio, Y.
\newblock 2015.
\newblock Show, attend and tell: Neural image caption generation with visual
  attention.
\newblock In {\em International conference on machine learning},  2048--2057.

\bibitem[\protect\citeauthoryear{Yang \bgroup et al\mbox.\egroup
  }{2018}]{yang2018mean}
Yang, Y.; Luo, R.; Li, M.; Zhou, M.; Zhang, W.; and Wang, J.
\newblock 2018.
\newblock Mean field multi-agent reinforcement learning.
\newblock In {\em Proceedings of the 35th International Conference on Machine
  Learning},  5567--5576.

\bibitem[\protect\citeauthoryear{Yu \bgroup et al\mbox.\egroup
  }{2015}]{yu2015multiagent}
Yu, C.; Zhang, M.; Ren, F.; and Tan, G.
\newblock 2015.
\newblock Multiagent learning of coordination in loosely coupled multiagent
  systems.
\newblock {\em IEEE Transactions on Cybernetics} 45(12):2853--2867.

\end{thebibliography}

\end{document}